\documentclass[10pt,twocolumn,letterpaper]{article}

\usepackage[pagenumbers]{cvpr} 

\usepackage{booktabs}
\usepackage{multirow}

\usepackage[dvipsnames]{xcolor}

\definecolor{cvprblue}{rgb}{0.21,0.49,0.74}
\usepackage[pagebackref,breaklinks,colorlinks,citecolor=cvprblue]{hyperref}

\def\paperID{24} 
\def\confName{CVPR}
\def\confYear{2024}

\title{Anytime, Anywhere, Anyone: Investigating the Feasibility of Segment Anything Model for Crowd-Sourcing Medical Image Annotations}

\author{Pranav Kulkarni\thanks{Authors contributed equally to this work.} \quad\quad
Adway Kanhere\textsuperscript{$*$} \quad\quad
Dharmam Savani\textsuperscript{$*$} \quad\quad
Andrew Chan \quad\quad \\
Devina Chatterjee \quad\quad
Paul H. Yi \quad\quad 
Vishwa S. Parekh\thanks{Corresponding author.} \\
University of Maryland Medical Intelligent Imaging (UM2ii) Center \\
University of Maryland School of Medicine, Baltimore, MD 21201 \\
{\tt\small \{pkulkarni,akanhere,dsavani,andrew.chan,devinachatterjee,pyi,vparekh\}@som.umaryland.edu}
}

\begin{document}
\maketitle

\begin{abstract}
Curating annotations for medical image segmentation is a labor-intensive and time-consuming task that requires domain expertise, resulting in "narrowly" focused deep learning (DL) models with limited translational utility. Recently, foundation models like the Segment Anything Model (SAM) have revolutionized semantic segmentation with exceptional zero-shot generalizability across various domains, including medical imaging, and hold a lot of promise for streamlining the annotation process. However, SAM has yet to be evaluated in a crowd-sourced setting to curate annotations for training 3D DL segmentation models. In this work, we explore the potential of SAM for crowd-sourcing "sparse" annotations from non-experts to generate "dense" segmentation masks for training 3D nnU-Net models, a state-of-the-art DL segmentation model. Our results indicate that while SAM-generated annotations exhibit high mean Dice scores compared to ground-truth annotations, nnU-Net models trained on SAM-generated annotations perform significantly worse than nnU-Net models trained on ground-truth annotations ($p<0.001$, all).
\end{abstract}

\section{Introduction}

Medical image segmentation is one of the most fundamental tasks in computer-assisted clinical decision support, forming the basis for many downstream applications from diagnosis to treatment planning and response assessment. However, developing medical image segmentation models requires a domain expert (e.g., a radiologist) to manually annotate different objects of interest across a training dataset consisting of hundreds of patients, making it an extremely labor-intensive and time-consuming task \cite{diaz2022monai,sebro2023totalsegmentator,wasserthal2023totalsegmentator}. As a result, most datasets and segmentation models developed in prior literature are "narrowly" focused only on the task at hand, thereby reducing their clinical translational utility. 

\begin{figure}[!t]
    \centering
    \includegraphics[width = \linewidth]{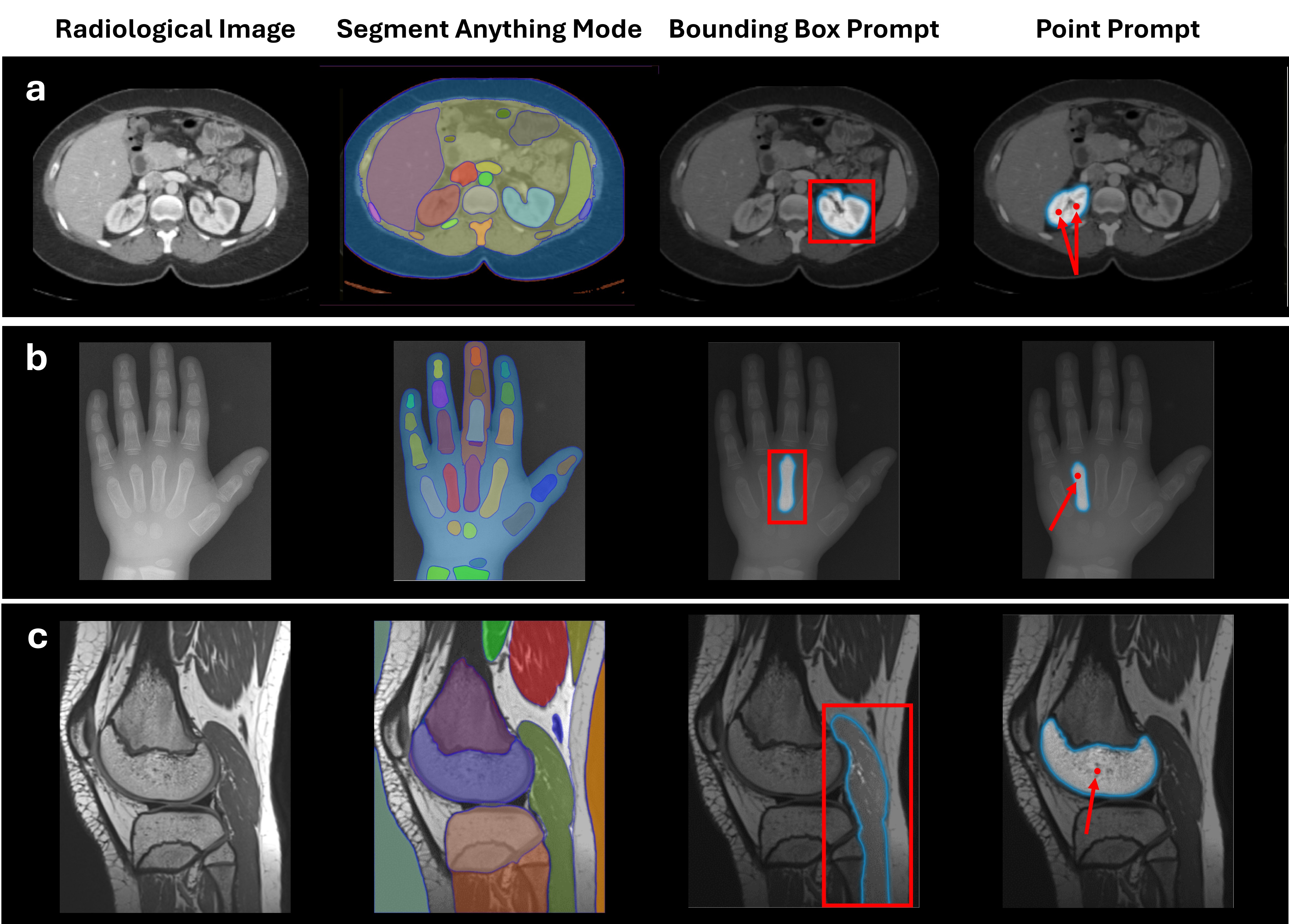}
    \caption{Example of SAM on \textbf{(a)} abdominal CT, \textbf{(b)} hand x-ray, and \textbf{(c)} knee MRI. SAM can operate in either "segment anything" mode (column 2), where SAM automatically segments all potential objects of interest in an image, or "prompting" mode, where SAM can segment an object of interest using an interactive prompt like bounding boxes (column 3) or points (column 4).}
    \label{fig:sam_overview}
\end{figure}

To address this challenge, many different approaches have been proposed in the recent years where users can use less time-consuming "sparse" annotations, such as scribbles and bounding boxes, to interactively prompt a pre-trained deep learning (DL) model to create "dense" annotations like detailed boundary masks \cite{diaz2022monai,ronneberger2015u,huang2018weakly}. While these approaches have shown to significantly reduce the annotation time per object, they require an expert to not just interactively create these annotations, but also fine-tune and validate them \cite{diaz2022monai}. Therefore, there is a critical need for a data curation pipeline for medical image segmentation that would allow non-experts to annotate datasets with sparse annotations without the need for an expert in the loop. 

\begin{table*}[!ht]
    \centering
    \caption{Dataset description and availability of organ annotations for the MSD Liver, MSD Spleen, and BTCV datasets.}
    \label{tab:dataset}
    \small
    \begin{tabular*}{\linewidth}{@{\extracolsep{\fill}} lccc} \toprule
        \textbf{Variable} & \multicolumn{1}{c}{\textbf{MSD Liver}} & \multicolumn{1}{c}{\textbf{MSD Spleen}} & \multicolumn{1}{c}{\textbf{BTCV}} \\ \midrule
        No. of Scans & 201 & 61 & 50 \\
        Modality & CT & CT & CT \\
        Volume Size (voxels) & 512 x 212 x 25 -- 512 x 212 x 287 & 512 x 212 x 25 -- 512 x 212 x 268 & 512 x 212 x 25 -- 512 x 212 x 295 \\
        In-Plane Resolution (mm\textsuperscript{2}) & 0.63 x 3.63 -- 1 x 1 & 0.73 x 3.73 -- 0.98 x 8.98 & 0.59 x 9.59 -- 0.98 x 8.98 \\
       Slice Thickness (mm) & 0.70 -- 5 & 1.5 -- 8 & 2.5 -- 5 \\ \midrule
       Aorta & - & - & 30 (60\%) \\
       Left Kidney & - & - & 30 (60\%) \\
       Right & - & - & 30 (60\%) \\
       Liver & 131 (65\%) & - & 30 (60\%) \\
       Spleen & - & 41 (67\%) & 30 (60\%) \\ \bottomrule
    \end{tabular*}
\end{table*}

Recently, large DL foundation models trained in a self-supervised manner on large-scale datasets on the order of a billion samples are revolutionizing the field of computer vision with their strong zero-shot generalizability \cite{kirillov2023segment,ma2024segment,butoi2023universeg}. This means that they do not need to be fine-tuned for medical imaging tasks and can operate directly out-of-the-box. The Segment Anything Model (SAM) is one such popular and open-source foundational model based on vision transformers (ViTs) with the ability for zero-shot semantic segmentation \cite{kirillov2023segment,dosovitskiy2020image}. SAM works by interactively prompting images with sparse annotations, such as points or bounding boxes, to generate segmentation masks (Figure \ref{fig:sam_overview}). 

Recent literature has suggested that SAM holds a lot of promise for annotating medical imaging datasets using sparse annotations \cite{cheng2023sam,bui2023sam3d,quan2024slide,deng2023sam,mazurowski2023segment,ma2024segment}. However, current approaches are limited to the evaluation of SAM in simulated settings as opposed to a realistic crowd-sourced setting and have yet to evaluate the effectiveness of SAM-generated annotations for training DL segmentation models. The purpose of this study is to 1) evaluate SAM for crowd-sourcing annotations for medical imaging datasets from non-expert annotators, and 2) investigate the feasibility of using SAM-generated annotations for training 3D DL segmentation models.

\section{Methods}

\begin{figure*}[!ht]
    \centering
    \includegraphics[width = \linewidth]{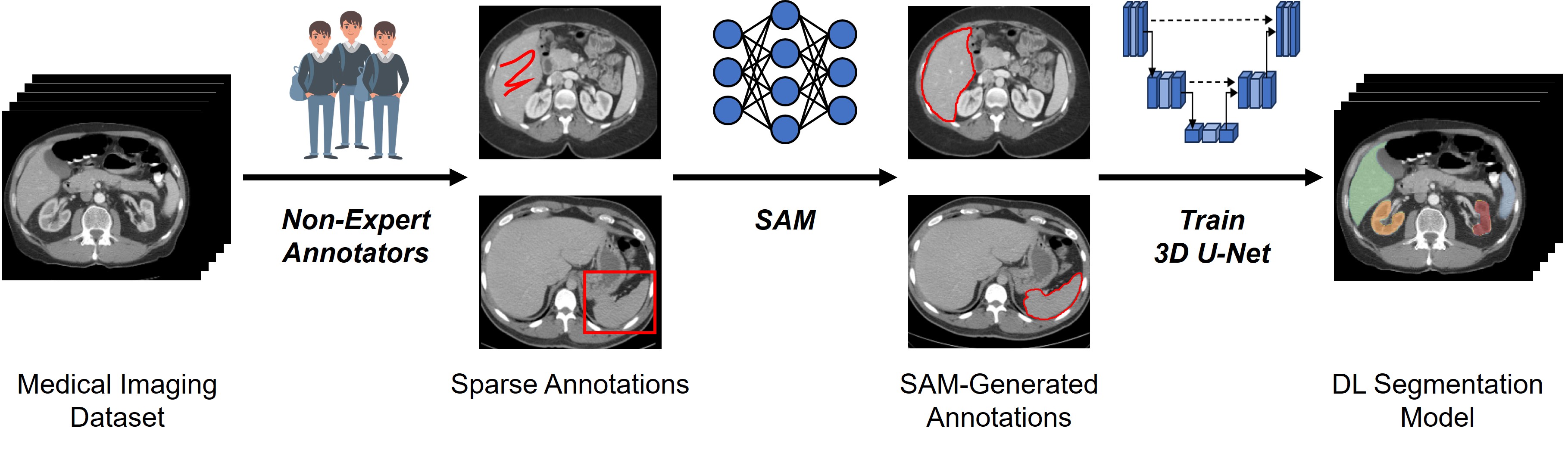}
    \caption{Pipeline for crowd-sourcing sparse annotations from non-expert annotators for the purpose of training 3D DL segmentation models using SAM-generated annotations. Suppose there is an unannotated medical imaging dataset. Sparse annotations for objects of interest (e.g., organs, tumors, etc.) can be crowd-sourced from non-expert annotators. Then, segmentation masks for the objects of interest can be generated using SAM. Finally, the SAM-generated annotations can be used to train a 3D DL segmentation model (e.g., U-Net).}
    \label{fig:pipeline_overview}
\end{figure*}

This retrospective study used publicly available datasets and was acknowledged by our IRB as non-human subjects research. Our code is available at: \url{https://github.com/UM2ii/SAM_DataAnnotation}

\subsection{Segment Anything Model}

The Segment Anything Model (SAM) is a computer vision foundation model for semantic segmentation based on ViTs \cite{kirillov2023segment,dosovitskiy2020image}. It is comprised of an image encoder that extracts features from image to create embeddings, a prompt encoder that creates embeddings from "sparse" annotations (e.g., points, bounding boxes), and a mask decoder that uses image and prompt embeddings to generate "dense" segmentation masks for the objects of interest in the image. SAM is trained on a large-scale dataset comprising of over 11 million images with over 1 billion segmentation masks. This enables SAM to have exceptional zero-shot and few-shot generalizability for semantic segmentation tasks across various domains, including medical imaging \cite{cheng2023sam,bui2023sam3d,deng2023sam,mazurowski2023segment,ma2024segment}.

\subsection{Datasets}

The Medical Segmentation Decathlon (MSD) is a collection of 10 benchmark datasets for segmentation spanning different body parts and modalities \cite{antonelli2022medical}. In our study, we used two datasets: 1) The MSD Liver dataset consists of $n=201$ portal-venous phase contrast-enhanced abdominal CT scans, out of which $n=131$ (65\%) contained annotations for the liver and liver tumors. We discarded all liver tumor annotations. 2) The MSD Spleen dataset consists of $n=61$ portal-venous phase contrast-enhanced abdominal CT scans, out of which $n=41$ (67\%) contained annotations for the spleen. The Beyond the Cranial Vault (BTCV) dataset consists of $n=50$ portal-venous phase contrast-enhanced abdominal CT scans, out of which $n=30$ (60\%) contained annotations for 13 abdominal organs \cite{landman2015miccai}. In our study, five organs of interest were included: aorta, left and right kidneys, liver, and spleen. The dataset description and availability of organ annotations for the MSD Liver, MSD Spleen, and BTCV datasets is provided in Table \ref{tab:dataset}.


\subsection{Crowd-Sourcing Sparse Annotations}

We implemented a pipeline for crowd-sourcing annotations with SAM for the purpose of training 3D DL segmentation models (Figure \ref{fig:pipeline_overview}). Since the volumes are provided in the NIfTI format, we used med2image (version 2.6.6) to convert a NIfTI volume into PNG images for each slice. The PNG format was chosen as it is a lossless standard format for image analysis in computer vision. We used the OpenLabeling tool (version 1.3) to annotate the organs of interest for each slice within a volume \cite{cartucho2018robust}. Bounding boxes were chosen as the sparse annotation method due to their superior performance for generating segmentation masks with SAM \cite{cheng2023sam,deng2023sam,mazurowski2023segment,ma2024segment}. In this study, we evaluate the effectiveness of SAM-generated annotations for training 3D DL segmentation models across two experiments: 1) simulated "sparse" annotations, and 2) crowd-sourced "sparse" annotations.

\begin{figure}[!t]
    \centering
    \includegraphics[width = \linewidth]{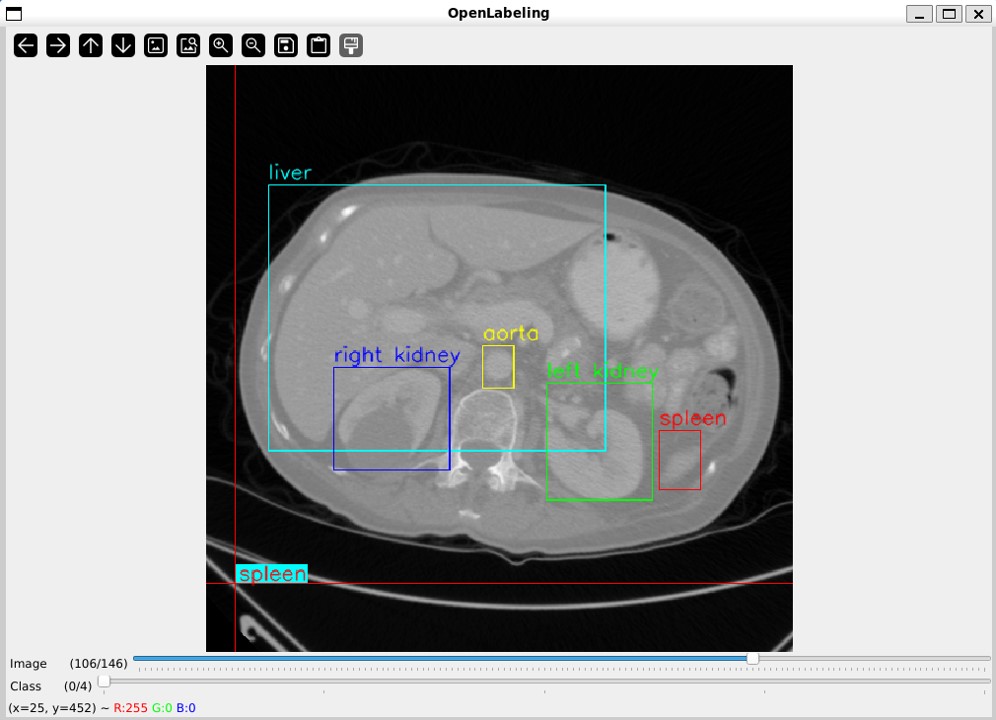}
    \caption{Illustration of the OpenLabeling tool used for crowd-sourcing bounding box annotations for the BTCV training set across the five organs of interest.}
    \label{fig:openlabeling_overview}
\end{figure}

\subsubsection{Simulated Annotations}

The goal of this experiment was to represent an "ideal" scenario for crowd-sourcing annotations, where all organs of interest are accurately annotated for every volume in the dataset. For liver segmentation, we use the MSD Liver and BTCV datasets as the training ($n=131$) and testing ($n=30$) sets respectively. Using the ground-truth annotations, the bounding box prompts were simulated for each slice of a volume in the training set. Ground-truth annotations were blurred using uniform filter (kernel size = 5 x 5) and binarized (thresholded to pixel values greater than 0.01) to introduce jitter. Bounding boxes were drawn programmatically around each resulting region of interest. For spleen segmentation, this process was repeated with the MSD Spleen dataset ($n=41$) as the training set.

\subsubsection{Crowd-Sourced Annotations}

The goal of this experiment was to represent a real-world scenario when crowd-sourcing annotations from non-experts. We randomly split the BTCV dataset into training ($n=15$) and testing ($n=15$) sets. The training set was annotated with the five organs of interest by four non-experts from diverse backgrounds with small-to-moderate knowledge of anatomical structures (Figure \ref{fig:openlabeling_overview}). Two annotators (A.K., D.S.) had software engineering backgrounds, one annotator (A.C.) had a bioengineering background, and one annotator (D.C.) was a medical student. All annotators had experience in biomedical informatics with a mean 2.50 years of experience (range 1--5 years). They were provided with basic orientation to familiarize them with relevant anatomical structures. Three annotators were assigned one task (D.S., aorta; A.C., liver; A.K., spleen) and one annotator was assigned two tasks (D.C., left and right kidneys). They were instructed to draw bounding boxes surrounding the region of interest for each slice in a volume. The volumes were annotated by the non-expert annotators independently, and consensus agreement was not required.

\subsection{Generating Dense Annotations}

After crowd-sourcing sparse annotations, we used SAM with the ViT-Huge backbone to generate segmentation masks for the organs of interest \cite{kirillov2023segment}. We also generated masks using MedSAM, a version of SAM fine-tuned for medical image segmentation tasks and based on the ViT-Base backbone \cite{ma2024segment}. Since SAM only supports 2D segmentation, annotations for a volume were generated by passing each slice with its corresponding bounding boxes as input to SAM for inference. The pixel values of all input images were normalized to range 0--255. The annotations were selected from three generated masks based on the highest confidence score. The SAM-generated annotations were converted to NIfTI using NiBabel (version 5.2). We measured the mean Dice similarity coefficient of SAM-generated annotations on the ground-truth annotations of the training set for the volume (hereafter, 'volume Dice score') and each annotated slice (hereafter, 'slice Dice score'). In all cases, the mean Dice scores are reported with the 95\% confidence interval as 'Mean $\pm$ 95\% CI'.

\subsection{Training DL Segmentation Models}

We train DL segmentation models using the nnU-Net framework on the SAM-generated (hereafter, 'SAM-nnU-Net') and ground-truth annotations (hereafter, 'GT-nnU-Net'). It is a self-configuring 3D U-Net architecture that automatically optimizes hyperparameters for the training dataset and has achieved state-of-the-art (SOTA) performance across various biomedical segmentation tasks \cite{isensee2021nnu,sebro2023totalsegmentator,wasserthal2023totalsegmentator,ronneberger2015u}. The models were trained with five-fold cross-validation for 1000 epochs using nnU-Net's default training procedure \cite{isensee2021nnu}. The mirroring augmentation was removed as it is not anatomically valid for abdominal CT scans \cite{wasserthal2023totalsegmentator}. We compared their performance using mean volume Dice scores on the ground-truth test set. All models were evaluated using PyTorch (version 2.0.1), and CUDA (version 12.0) on eight NVIDIA GeForce RTX A6000 GPUs.

\subsection{Statistical Analysis}

The mean Dice scores were compared with Wilcoxon signed-rank tests due to the non-parametric distribution of paired samples, as indicated by $p < 0.05$ using the Shapiro-Wilk test for normality. All statistical analysis was performed by a statistical analyst (P.K., 6 years of experience) using SciPy (version 1.11.2). Statistical significance was defined as $p < 0.05$.

\section{Results}

\subsection{Simulated Annotations}

\subsubsection{Liver Segmentation}

The SAM-generated annotations measured a mean volume Dice score of $0.86 \pm 0.02$ on the ground-truth annotations for the MSD Liver training set, which is significantly higher than the mean volume Dice score of 0$.80 \pm 0.04$ ($p < 0.001$) for MedSAM. Similarly, the SAM-generated annotations measured a mean slice Dice score of $0.85 \pm 0.00$ on the ground-truth annotations for the MSD Liver training set, which is significantly higher than the mean slice Dice score of $0.77 \pm 0.00$ ($p < 0.001$) for MedSAM. The SAM-nnU-Net model trained on the MSD Liver dataset measured a mean volume Dice score of $0.83 \pm 0.03$ on the BTCV dataset, which is significantly lower than the mean volume Dice score of $0.90 \pm 0.03$ ($p < 0.001$) for GT-nnU-Net. 

\subsubsection{Spleen Segmentation}

The SAM-generated annotations measured a mean volume Dice score of $0.88 \pm 0.02$ on the ground-truth annotations for the MSD Spleen training set, which is comparable to the mean volume Dice score of $0.87 \pm 0.03$ ($p = 0.23$) for MedSAM. Similarly, the SAM-generated annotations measured a mean slice Dice score of $0.87 \pm 0.01$ on the ground-truth annotations for the MSD Spleen training set, which is significantly higher than the mean slice Dice score of $0.83 \pm 0.01$ ($p < 0.001$) for MedSAM. The SAM-nnU-Net model trained on the MSD Spleen dataset measured a mean volume Dice score of $0.81 \pm 0.01$ on the BTCV dataset, which is significantly lower than the mean volume Dice score of $0.87 \pm 0.01$ ($p = 0.004$) for GT-nnU-Net. 

\subsection{Crowd-Sourced Annotations}

\begin{figure*}[!ht]
    \centering
    \includegraphics[width = \linewidth]{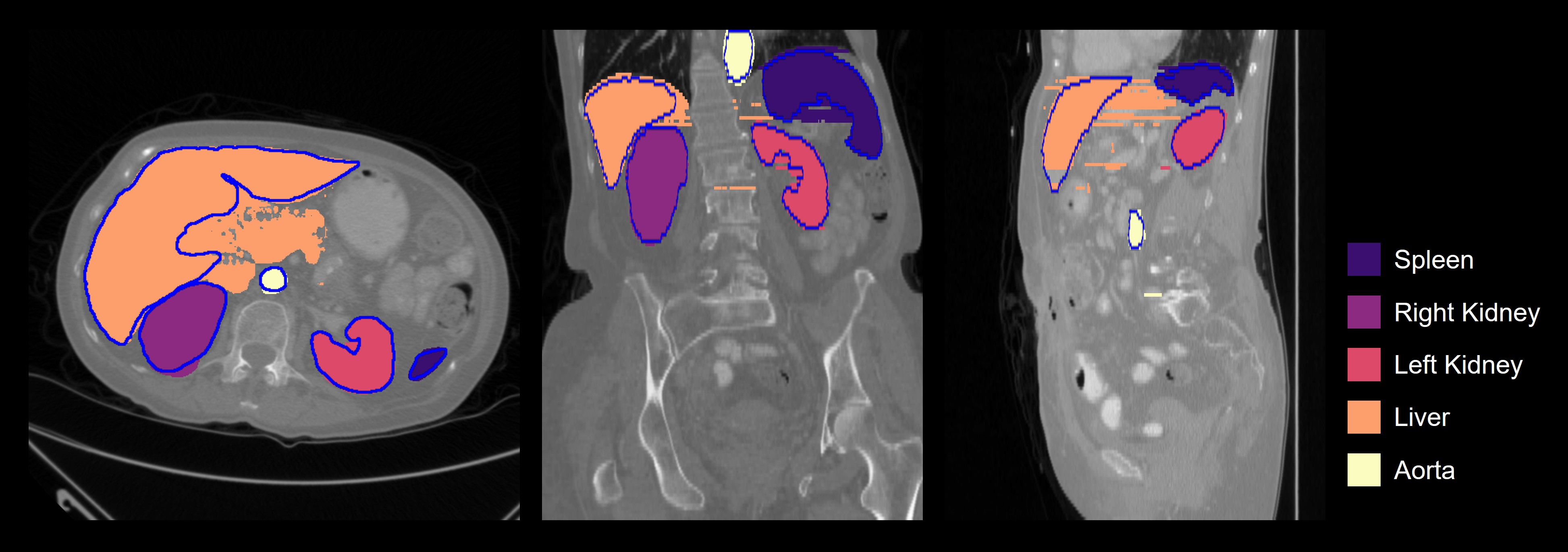}
    \caption{An example of crowd-sourced SAM-generated annotations for a CT scan from the BTCV training set in the axial, coronal, and sagittal views. The SAM-generated annotations are filled in while the ground-truth annotations are outlined in blue.}
    \label{fig:btcv_example}
\end{figure*}

The non-expert annotators annotated 651 slices from $n=15$ volumes in the BTCV training set with 1840 bounding boxes using our data curation pipeline (Figure \ref{fig:btcv_example}). They took $55.60 \pm 8.76$ mins to annotate an organ of interest across all volumes, with a mean of $3.29 \pm 1.04$ seconds per slice. Out of $n=15$ volumes, the non-experts annotated all five organs of interest for $n=11$ volumes (73\%). For the remaining $n=4$ volumes (27\%), annotations for left and right kidneys were missing for $n=3$ volumes (75\%), liver was missing for $n=2$ volumes (50\%), and spleen was missing for $n=1$ volume (25\%).

For measuring the mean volume and slice Dice scores for SAM- and MedSAM-generated annotations, we exclude the $n=4$ volumes with missing annotations. The SAM-generated annotations measured a mean volume Dice score of $0.75 \pm 0.09$ on the ground-truth annotations for the BTCV training set, compared to the mean volume Dice score of $0.74 \pm 0.09$ ($p = 0.70$) for MedSAM. Stratified by organs of interest, the SAM-generated annotations were comparable to MedSAM-generated annotations for all five organs ($p \geq 0.05$, all) (Table \ref{tab:btcv_volume_dice}). Similarly, the SAM-generated annotations measured a mean slice Dice score of $0.88 \pm 0.02$ on the ground-truth annotations for the BTCV training set, compared to the mean volume Dice score of $0.88 \pm 0.02$ ($p = 0.34$) for MedSAM. Stratified by organs of interest, the SAM-generated annotations were comparable to MedSAM-generated annotations for left and right kidneys ($p \geq 0.05$, both) (Table \ref{tab:btcv_slice_dice}). However, the mean slice Dice scores for MedSAM-generated annotations were significantly higher than the mean slices Dice scores for SAM-generated annotations for aorta ($0.89 \pm 0.02$ vs $0.88 \pm 0.01$, $p = 0.02$) and liver ($0.89 \pm 0.02$ vs $0.84 \pm 0.02$, $p < 0.001$), but significantly lower for spleen ($0.88 \pm 0.04$ vs $0.91 \pm 0.03$, $p = 0.02$).

\begin{table}[!t]
    \centering
    \caption{Mean volume Dice scores for the SAM- and MedSAM-generated annotations on the ground-truth annotations of the BTCV training set ($n=15$) across all five organs of interest.}
    \label{tab:btcv_volume_dice}
    \small
    \begin{tabular*}{\linewidth}{@{\extracolsep{\fill}} llll} \toprule
        \textbf{Organ} & \multicolumn{1}{c}{\textbf{SAM}} & \multicolumn{1}{c}{\textbf{MedSAM}} & \multicolumn{1}{c}{\textbf{p-value}} \\ \midrule
        Aorta & $\mathbf{0.70} \pm 0.09$ & $0.65 \pm 0.09$ & $0.08$ \\
        Left Kidney & $\mathbf{0.74} \pm 0.11$ & $0.74 \pm 0.10$ & $0.77$ \\
        Right Kidney & $\mathbf{0.78} \pm 0.10$ & $0.77 \pm 0.09$ & $0.97$ \\
        Liver & $0.73 \pm 0.13$ & $\mathbf{0.75} \pm 0.14$ & $0.12$ \\
        Spleen & $\mathbf{0.80} \pm 0.14$ & $0.78 \pm 0.14$ & $0.90$ \\ \bottomrule
    \end{tabular*}
\end{table}

We trained two sets of 3D DL segmentation models: 1) Trained on partially annotated $n=15$ volumes from the BTCV training set. 2) Trained on fully annotated $n=11$ volumes from the BTCV training set. Both sets of DL models were evaluated on the BTCV test set ($n=15$). 

The SAM-nnU-Net model trained on partially annotated $n=15$ volumes from the BTCV training set measured a mean Dice score of $0.77 \pm 0.06$ on the BTCV test set, which is significantly lower than the mean Dice score of $0.91 \pm 0.05$ ($p < 0.001$) for the GT-nnU-Net model. Stratified by organs of interest, the SAM-nnU-Net model had a significantly lower mean Dice score than the GT-nnU-Net model for all five organs ($p < 0.006$, all) (Table \ref{tab:btcv_nnunet_15}). 

\begin{figure*}[!ht]
    \centering
    \includegraphics[width = \linewidth]{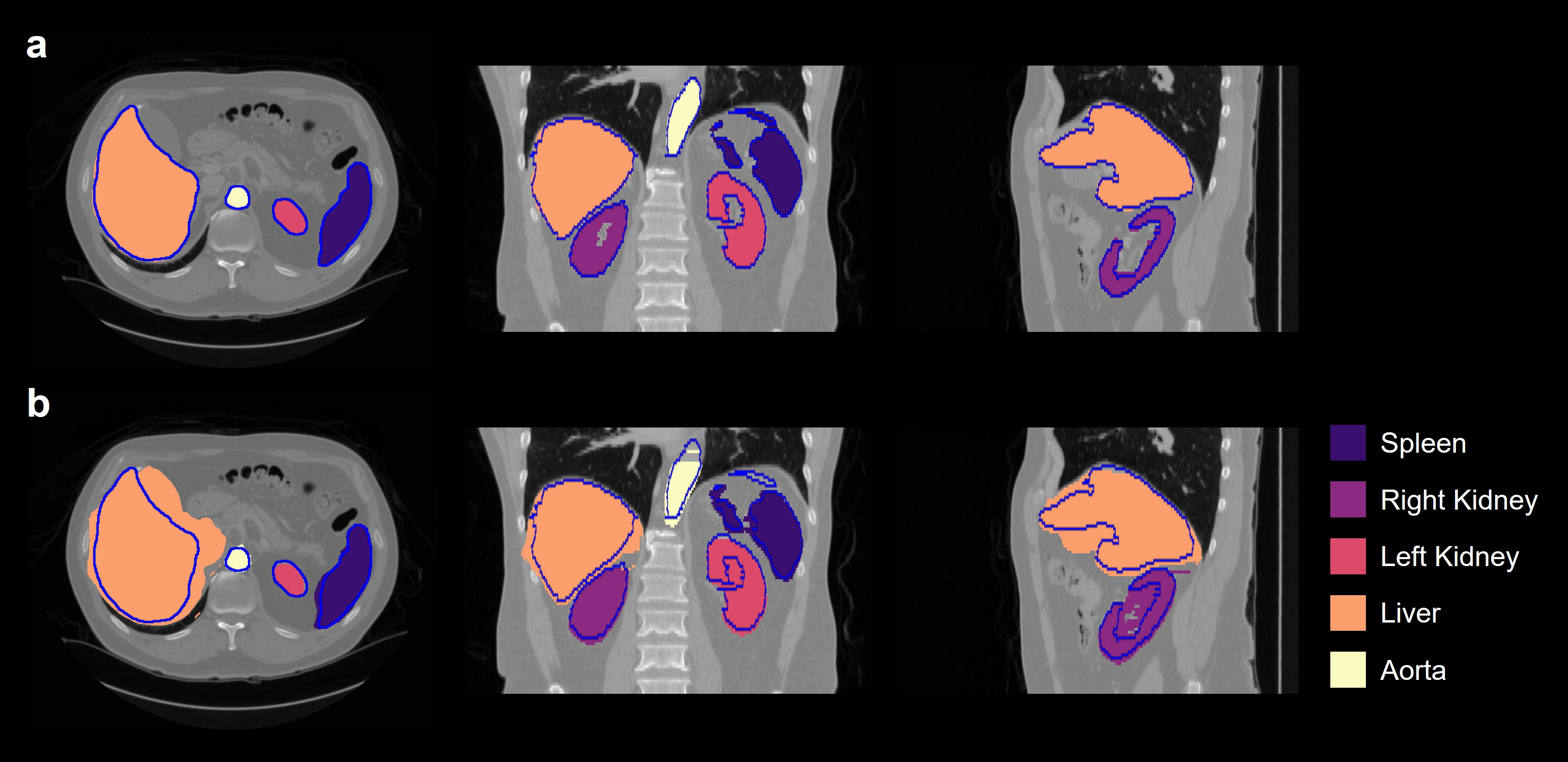}
    \caption{An example of \textbf{(a)} GT-nnU-Net and \textbf{(b)} SAM-nnU-Net segmentations for a CT scan from the BTCV test set in the axial, coronal, and sagittal views. The models are trained on fully annotated n=11 volumes from the BTCV training set. The predicted segmentations are filled in while the ground-truth annotations are outlined in blue.}
    \label{fig:nnunet_btcv_pred}
\end{figure*}

The SAM-nnU-Net model trained on fully annotated $n=11$ volumes from the BTCV training set measured a mean Dice score of $0.80 \pm 0.05$ on the BTCV test set, which is significantly lower than the mean Dice score of $0.90 \pm 0.05$ ($p < 0.001$) for the GT-nnU-Net model (Figure \ref{fig:nnunet_btcv_pred}). Stratified by organs of interest, the SAM-nnU-Net model had a significantly lower mean Dice score than the GT-nnU-Net model for aorta, left kidney, liver, and spleen ($p < 0.02$, all) (Table \ref{tab:btcv_nnunet_11}). However, for right kidneys, the mean Dice score of $0.78 \pm 0.07$ for the SAM-nnU-Net model was comparable to the mean Dice score of $0.87 \pm 0.11$ for the GT-nnU-Net model ($p = 0.06$).

\begin{table}[!t]
    \centering
    \caption{Mean slice Dice scores for the SAM- and MedSAM-generated annotations on the ground-truth annotations of the BTCV training set ($n=15$) across all five organs of interest.}
    \label{tab:btcv_slice_dice}
    \small
    \begin{tabular*}{\linewidth}{@{\extracolsep{\fill}} llll} \toprule
        \textbf{Organ} & \multicolumn{1}{c}{\textbf{SAM}} & \multicolumn{1}{c}{\textbf{MedSAM}} & \multicolumn{1}{c}{\textbf{p-value}} \\ \midrule
        Aorta & $0.88 \pm 0.01$ & $\mathbf{0.89} \pm 0.02$ & $\underline{0.02}$ \\
        Left Kidney & $0.88 \pm 0.04$ & $\mathbf{0.89} \pm 0.02$ & $0.66$ \\
        Right Kidney & $\mathbf{0.90} \pm 0.02$ & $0.88 \pm 0.02$ & $0.11$ \\
        Liver & $0.84 \pm 0.02$ & $\mathbf{0.89} \pm 0.02$ & $\underline{<0.001}$ \\
        Spleen & $\mathbf{0.91} \pm 0.03$ & $0.86 \pm 0.04$ & $\underline{0.02}$ \\ \bottomrule
    \end{tabular*}
\end{table}

By excluding the $n=4$ volumes with missing annotations from the BTCV training set, we observe that a significant increase in mean Dice of the SAM-nnU-Net model from $0.77 \pm 0.06$ to $0.80 \pm 0.05$ ($p = 0.008$) (Tables \ref{tab:btcv_nnunet_15} and \ref{tab:btcv_nnunet_11}). Stratified by organs of interest, we observe significant improvements in mean Dice scores for aorta ($p = 0.04$), left kidney ($p = 0.002$), right kidney ($p = 0.02$), and liver ($p = 0.04$). However, for the GT-nnU-Net model, we observe no significant differences in mean Dice score after excluding volumes with missing annotations ($p = 0.06$) (Tables \ref{tab:btcv_nnunet_15} and \ref{tab:btcv_nnunet_11}). Stratified by organs of interest, we observe a significant decrease in mean Dice scores for aorta ($p = 0.008$) and right kidney ($p = 0.04$).

\begin{table}[!t]
    \centering
    \caption{Mean volume Dice scores of the GT-nnU-Net and SAM-nnU-Net on the BTCV test set ($n=15$) across all five organs of interest. The models are trained on partially annotated $n=15$ volumes from the BTCV training set.}
    \label{tab:btcv_nnunet_15}
    \small
    \begin{tabular*}{\linewidth}{@{\extracolsep{\fill}} llll} \toprule
        \textbf{Organ} & \multicolumn{1}{c}{\textbf{GT-nnU-Net}} & \multicolumn{1}{c}{\textbf{SAM-nnU-Net}} & \multicolumn{1}{c}{\textbf{p-value}} \\ \midrule
        Aorta & $\mathbf{0.93} \pm 0.01$ & $0.75 \pm 0.05$ & $\underline{<0.001}$ \\
        Left Kidney & $\mathbf{0.90} \pm 0.11$ & $0.72 \pm 0.11$ & $\underline{<0.001}$ \\
        Right Kidney & $\mathbf{0.90} \pm 0.10$ & $0.75 \pm 0.07$ & $\underline{0.005}$ \\
        Liver & $\mathbf{0.94} \pm 0.05$ & $0.82 \pm 0.05$ & $\underline{<0.001}$ \\
        Spleen & $\mathbf{0.87} \pm 0.12$ & $0.79 \pm 0.12$ & $\underline{<0.001}$ \\ \bottomrule
    \end{tabular*}
\end{table}

\section{Discussion}

Crowd-sourcing of annotations using foundation models, like SAM, has the potential to revolutionize the curation of large-scale datasets for medical image segmentation. It transforms a labor-intensive and time-consuming process, that requires domain expertise, into one that allows anyone, whether experts or non-experts, to annotate objects of interest in a medical image using "sparse" annotations from anywhere, using any device, and at any time without the need for an expert in the loop.

However, our results indicate that while SAM-generated annotations exhibit high mean slice Dice scores compared to ground-truth annotations, the SAM-nnU-Net models perform significantly worse than the GT-nnU-Net models across the simulated setting and the crowd-sourced setting for the segmentation of abdominal organs using CT scans ($p<0.001$, all). This discrepancy is due SAM being primarily designed for 2D semantic segmentation, thereby resulting in a lack of spacial relationships between features in 3D (e.g., depth) and poor "connectivity" in annotations between two consecutive slices (Figure \ref{fig:btcv_example}). As indicated by our results, this leads to significantly worse mean volume Dice scores for SAM-generated annotations when compared to ground-truth annotations and translates into sub-optimal performance for the SAM-nnU-Net models, despite nnU-Net being a SOTA model architecture.

\begin{table}[!t]
    \centering
    \caption{Mean volume Dice scores of the GT-nnU-Net and SAM-nnU-Net on the BTCV test set ($n=15$) across all five organs of interest. The models are trained on partially annotated $n=11$ volumes from the BTCV training set.}
    \label{tab:btcv_nnunet_11}
    \small
    \begin{tabular*}{\linewidth}{@{\extracolsep{\fill}} llll} \toprule
        \textbf{Organ} & \multicolumn{1}{c}{\textbf{GT-nnU-Net}} & \multicolumn{1}{c}{\textbf{SAM-nnU-Net}} & \multicolumn{1}{c}{\textbf{p-value}} \\ \midrule
        Aorta & $\mathbf{0.92} \pm 0.01$ & $0.78 \pm 0.04$ & $\underline{<0.001}$ \\
        Left Kidney & $\mathbf{0.87} \pm 0.12$ & $0.78 \pm 0.08$ & $\underline{0.02}$ \\
        Right Kidney & $\mathbf{0.87} \pm 0.11$ & $0.78 \pm 0.07$ & $0.06$ \\
        Liver & $\mathbf{0.94} \pm 0.05$ & $0.84 \pm 0.04$ & $\underline{<0.001}$ \\
        Spleen & $\mathbf{0.88} \pm 0.11$ & $0.80 \pm 0.11$ & $\underline{<0.001}$ \\ \bottomrule
    \end{tabular*}
\end{table}

The disparity can be addressed by developing foundation models specialized for 3D semantic segmentation that are able to capture spacial relationships between features -- a key difference between natural images and medical images. Recent literature has explored 3D adapters that retain SAM's rich knowledge base for semantic segmentation while incorporating spacial relationships when generating "dense" segmentation masks in 3D \cite{lei2023medlsam,quan2024slide,wu2023medical}. One popular approach is using a 3D CNN decoder to capture spacial relationships between image embeddings created slice-by-slice by SAM's image encoder \cite{bui2023sam3d,gong20233dsam}. These variations have demonstrated significantly better performance in 3D medical images when compared to SAM and MedSAM.

Another crucial consideration for crowd-sourcing annotations is that there is a potential for unreliable and spurious annotations due to the uncertainty associated with non-expert annotators. Our results indicate the importance of quality assessment. In our study, the non-expert annotators failed to completely annotate $n=4$ volumes (27\%) with all five organs of interest, resulting in a partially annotated and unreliable dataset for training DL segmentation models. By filtering out incomplete and unreliable annotations, we observed a significant increase in the performance of the SAM-nnU-Net model.

Therefore, robust quality assessment is critical for not just evaluating the quality of crowd-sourced annotations, but also filtering out low-quality annotations without manual intervention from an expert. One potential technique is inter-rater reliability where low-quality annotations with poor agreement with other annotations can be identified and filtered out from the study without impacting the quality of annotations generated by more reliable annotators. Moreover, this enables unreliable and spurious annotators with consistently low-quality annotations to be similarly identified and removed from the study. In addition, uncertainty estimation of SAM-generated annotations using multiple bounding box prompts can be used identify potential errors in SAM-generated annotations and guide the selection of higher-quality annotations \cite{deng2023sam}.

Our work has certain limitations. 1) We only consider SAM for our study, which is primarily designed for 2D segmentation with natural images and lacks the spacial relationships critical for 3D segmentation with medical images. 2) While our simulated experiments encapsulate large-scale datasets, we use a small dataset ($n=15$) for crowd-sourcing "sparse" annotations from non-experts to train DL segmentation models. 3) In our crowd-sourcing experiment, each annotator is assigned one organ of interest. This has the potential for unreliable and spurious annotations due to the absence of consensus agreements with multiple annotators. 4) We only consider CT scans and have not evaluated the potential for crowd-sourcing for other modalities like MRI and PET scans.

For future work, we intend to expand the scale of our study with larger multi-organ segmentation dataset than the BTCV dataset, larger number of expert and non-expert annotators for crowd-sourcing "sparse" annotations, and specialized adapters of SAM for 3D medical image segmentation. We also intend to include quality assessment metrics like inter-rater reliability to filter out unreliable and spurious annotations while retaining high-quality annotations for our analysis.

\section{Conclusion}

Limitations in current approaches warrant caution before incorporating crowd-sourced annotations from non-experts. To take full advantage of crowd-sourcing, specialized foundation models need to be developed for 3D segmentation. Furthermore, crowd-sourcing approaches should incorporate quality assessment to filter out low-quality annotations. While we may not be ready for non-expert annotations yet, they have the potential for streamlining the annotation process for medical image segmentation by enabling anyone to annotate medical images from anywhere and at any time.

{
    \small
    \bibliographystyle{ieeenat_fullname}
    \bibliography{main}
}


\end{document}